\documentclass{article}

\usepackage[preprint]{neurips_2025}

\usepackage[utf8]{inputenc}
\usepackage[T1]{fontenc}
\usepackage{hyperref}
\usepackage{xurl}
\usepackage{booktabs}
\usepackage{amsfonts}
\usepackage{nicefrac}
\usepackage{microtype}
\usepackage{xcolor}
\usepackage{amsmath} 
\usepackage[colorinlistoftodos]{todonotes}
\usepackage{multirow} 
\usepackage{tikz}
\usepackage{enumitem}

\usepackage{graphicx}
\usepackage{subcaption}

\newcommand{\mupax}{\textsc{MuPAX}}
\newcommand{\fullmupax}{\textsc{Multidimensional Problem-Agnostic eXplainable AI (MuPAX)}}

\newcommand{\imsize}{0.157}
\newcommand{\N}[1]{\mathit{N_{#1}}}

\newcommand{\roundedimage}[2]{%
\begin{tikzpicture}
    \clip [rounded corners=8pt] (0,0) rectangle (#2, #2);
    \node [anchor=south west, inner sep=0] at (0,0)
        {\includegraphics[width=#2]{#1}};
\end{tikzpicture}%
}

\title{\mupax: Multidimensional Problem--Agnostic eXplainable AI}

\author{%
  Vincenzo Dentamaro\thanks{Corresponding author.}\\
  Department of Computer Science\\
  University of Bari “Aldo Moro”\\
  Via E. Orabona 4, 70125 Bari, Italy\\
  \texttt{vincenzo.dentamaro@uniba.it}\\
  \And
  Felice Franchini\\
  Department of Computer Science\\
  University of Bari “Aldo Moro”\\
  Via E. Orabona 4, 70125 Bari, Italy\\
  \texttt{f.franchini9@phd.uniba.it}\\
  \And
  Giuseppe Pirlo\\
  Department of Computer Science\\
  University of Bari “Aldo Moro”\\
  Via E. Orabona 4, 70125 Bari, Italy\\
  \texttt{giuseppe.pirlo@uniba.it}\\
  \And
  Irina Voiculescu\\
  Department of Computer Science\\
  University of Oxford\\
  Wolfson Building, Parks Rd, Oxford OX1 3QG, United Kingdom\\
  \texttt{irina@cs.ox.ac.uk}\\
}

\begin{document}

\maketitle

\begin{abstract}

Robust XAI techniques should ideally be simultaneously  deterministic, model-agnostic, and guaranteed to converge. We propose \fullmupax, a deterministic, model-agnostic  explainability technique, with guaranteed convergency. \mupax's measure-theoretic formulation gives principled feature importance attribution through structured perturbation analysis that discovers inherent input patterns and eliminates spurious relationships. We  evaluate \mupax\ on an extensive range of data modalities and tasks: audio classification (1D), image classification (2D), volumetric medical image analysis (3D), and anatomical landmark detection, demonstrating dimension-agnostic effectiveness. The rigorous convergence guarantees extend to any loss function and arbitrary dimensions, making \mupax\ applicable to virtually any problem context for AI. By contrast with other XAI methods that typically decrease performance when masking,  \mupax\ not only preserves but actually enhances model accuracy by capturing only the most important patterns of the original data.
Extensive benchmarking against the state of the XAI art demonstrates \mupax's ability to generate precise, consistent and understandable explanations, a crucial step towards explainable and trustworthy AI systems.
The source code will be released upon publication.

\end{abstract}
\section{Introduction}
\label{sec:intro}

Explainable AI  (XAI) techniques suffer from well-known shortcomings. Perturbation-based strategies like LIME \cite{ribeiro2016should} has been shown to be affected by run-to-run instability \cite{visani2022statistical}. Gradient-based tools like Grad-CAM \cite{selvaraju2017grad} are not applicable to generic model architectures \cite{buhrmester2021analysis}. 
Frameworks like SHAP \cite{lundberg2017unified}, developed with a robust theoretical background, can be hindered  by computational issues in higher-dimensional spaces \cite{van2022tractability}. Such shortcomings point towards a need for XAI techniques which are simultaneously {\em deterministic, model-agnostic, and guaranteed to converge}.

\fullmupax\ is a novel, problem-agnostic and signal-agnostic methodology to perform explainable AI that works across any ML problem domain. We introduce a theoretical framework that operates in N-dimensions with arbitrary loss functions, supported by formal convergence guarantees. We validate \mupax\ across diverse data modalities and for different tasks (image classification, volumetric medical image analysis, anatomical landmark detection). \mupax\ is a novel XAI paradigm rooted in measure-theoretic convergence principles that gives a formal definition of feature importance through the \mupax\ Theorem. The \mupax\ Theorem provides dimension-agnostic convergence guarantees for any AI model working on structured  higher dimension data volumes. As a side-effect of 
using only the most important patterns extracted from the original data signal, \mupax\ not only returns explanations but also enhances the model's performance.

The output that \mupax\ generates is a saliency map containing only the most important information in the original signal weighted by their importance as a result of the structured perturbation sampling and loss choice. For example, in the case of 1D audio, are the most important frequencies in time; for 2D images are the pixels that retain the most of the information; for 3D volumetric data, it retains voxels that contain the majority of informative content; and for landmark detection,  are the image regions that lead to the correct localization of each anatomical keypoint.

\mupax's deterministic explanations directly serve upcoming AI laws like the EU AI Act, which demands traceability in safety-critical systems \cite{brkan2020legal}. In healthcare, the robutstness of our method enables radiologists to validate AI diagnosis on statistically relevant imaging areas rather than heuristic hotspots \cite{van2022medical}. 

The work is organized as follows: Section \ref{sec:related} presents related XAI work. Section \ref{sec:evidence} details the \mupax\ theorem. Experiments and results are in Section \ref{sec:experiments}, benchmarking different XAI algorithms. A more general discussion is in  Section \ref{sec:discussion}. Section \ref{sec:conclusion} concludes and comments on future work ideas.

\section{Related XAI works}
\label{sec:related}

The recent boom in XAI literature has led to progress in AI interpretability on multiple fronts.

\textbf{Decision Trees and Bayesian Networks:} Offer inherent interpretability through recursive partitioning of the input space or probabilistic modeling of variable dependencies \cite{blanco2019machine, derks2020taxonomy}.

\textbf{Shapley Additive Explanations (SHAP):} Founded on Shapley values in game theory, SHAP approximates feature contribution to single predictions, and its optimal application is on tabular data \cite{vandbroeck2022tractability}.
\textbf{Local Interpretable Model-Agnostic Explanations (LIME):} Interpolates black-box models with local neighborhoods of interpretable surrogates around instances \cite{visani2022statistical, hamilton2022enhancing}. 
Both SHAP and LIME are model-agnostic, in that they work on any model without using internal features.
\textbf{DLIME:} Solves LIME's randomness by using deterministic perturbation for more robust explanations \cite{zafar2021deterministic, shi2020modified, schlegel2022tsmule}. 
\textbf{Gradient-Based Methods:} Techniques like Integrated Gradients generate saliency maps that relate input changes to outputs, making attribution analysis easier \cite{bodria2021benchmarking, buhrmester2021analysis, machlev2022explainable}. These methods are global, in that they explain overall model behavior across an entire input space.
\textbf{Class Activation Maps (CAMs) and GradCAM:} Emphasize image areas that play a role in classification in CNNs, with GradCAM making use of gradient information \cite{byun2022recipro, joo2019visualization, marmolejosaucedo2022numerical}. These methods are model-spcific, in that they require internal information from the model. GradCAM is the de-facto technique used for landmark detection problem explainability \cite{li2019spatial, adewole2021deep}.
\textbf{Layer-wise Relevance Propagation and SmoothGrad:} LRP computes relevance scores of input features; SmoothGrad refines explanations via gradients \cite{jung2021explaining, ullah2021explaining}. 

 Depending on the design of each XAI mthod, different types of XAI functionality have emerged.

\textbf{Black Box Explainers} refer to  post-hoc methods which explain opaque models using gradients/perturbations. Examples of these are Gradients, Integrated Gradients, DeepLIFT \cite{bodria2021benchmarking, buhrmester2021analysis, machlev2022explainable, bhat2022gradient, vanderVelden2022explainable}.
\textbf{White Box Creators} are inherently interpretable models which output transparent decisions \cite{chou2022benchmark, joel2020explaining, brkan2020legal}.
\textbf{Equity Promoters} mitigate bias so as to ensure fair and equitable predictions. Examples of these are Fairness Constraints, Fair Representation Learning, Pre-/Post-processing Techniques \cite{calegari2020integration, li2023trustworthy, kaadoud2021explainable, charte2020analysis, qian2021xnpl, jain2020biased}.
\textbf{Sensitivity Analyzers} assess model robustness via input perturbation analysis. Examples are Local/Global Sensitivity Analysis, Perturbation-Based Methods \cite{yeh2019fidelity, molnar2021interpretable, wang2020perturbation, jhala2020perturbation, rashki2019system}.

\section{\mupax\ Theory}
\label{sec:evidence}

\paragraph{Setup.}
Let $\mathbf{X}\,\in\,\mathbb{R}^{\,d_1 \times \cdots \times d_N}$
be a nonnegative $N$-dimensional input (e.g.\ a multi-channel image). Consider a \emph{frozen} model
$$
f: \;\mathbb{R}^{\,d_1 \times \cdots \times d_N} \;\to\; \mathcal{Y},
$$
which could be a classifier, a regressor, a landmark-detection network, or a generative model. A user-chosen error (or loss) function
$$
L: \;\mathcal{Y}\times \mathcal{T}\;\to\;\mathbb{R}_{\ge 0}
$$
assesses the quality of the model output relative to a target $\mathbf{y}_{\text{true}} \in \mathcal{T}$, thereby defining
$$
\mu(\mathbf{X}_{\text{arg}}) 
\;=\; 
L\bigl(f(\mathbf{X}_{\text{arg}}),\, \mathbf{y}_{\text{true}}\bigr)
\;\in\;\mathbb{R}_{\ge 0}.
$$
Where $\mathbf{X}_{\text{arg}}$ is the input being fed to the model $f$. Lower values of $\mu(\mathbf{X}_{\text{arg}})$ imply the presence of “recognizable” patterns from the model (preserving essential features for the model), whereas higher values imply “misleading” or unhelpful patterns.

\paragraph{Chunking and Masking (N-Dim).}
Partition $\mathbf{X}$ into $m$ hyper-rectangular \emph{chunks}, each chunk being either retained $(1)$ or zeroed $(0)$. Denote $s\;\in\;\{0,1\}^m$ as the selection vector, and let $\mathbf{F}^s$  be the corresponding filter tensor with ones in the retained chunks. The \emph{filtered input} is 
$$
\mathbf{X}^s \;=\;\mathbf{X}\,\circ\,\mathbf{F}^s,
$$
and its associated error is $\mu_s = \mu(\mathbf{X}^s)$. An inverse-error weight is similarly defined by
$$
\mu'_s \;=\;\frac{1}{\,\mu_s + 1\,}.
$$

\paragraph{Goal.}
Given that $n$ masked samples $\{\overline{\mathbf{X}}^1,\dots,\overline{\mathbf{X}}^n\}$ produce error values below some threshold $W$ (or otherwise selected), it is possible to determine which coordinates of $\mathbf{X}$ consistently contribute to “good” or “bad” performance under the model $f$. 
In practice, generating the set $\{\overline{\mathbf{X}}^1,\dots,\overline{\mathbf{X}}^n\}$ involves randomly sampling a large number $N_{\text{total}} \gg n$ of selection vectors $s$, generating the corresponding filtered inputs $\mathbf{X}^s$, computing their error $\mu_s$, and retaining the first $n$ samples for which $\mu_s \le W$. This approximates drawing independent and identically distributed (i.i.d.) samples from the truncated distribution $\{\mathbf{X}_{\text{arg}} : \mu(\mathbf{X}_{\text{arg}}) \le W\}$, assuming $N_{\text{total}}$ is sufficiently large relative to the probability mass satisfying the condition. The theoretical results rely on the properties of this selected subset.

\subsection{\mupax\ General XAI Theorem}
\label{thm:evidence-general}

Let $\mathcal{U}$ be a base distribution over selection vectors $s \in \{0,1\}^m$ (in this work we used stratified uniform sampling).
Let $W > 0$ be a fixed threshold such that the acceptance probability $p_W = \mathbb{P}_{s \sim \mathcal{U}}(\mu(\mathbf{X}^s) \le W)$ is strictly positive. In practice, $W$ is set as the $20^{th}$ percentile of error values observed during an initial sampling phase from $\mathcal{U}$.

Samples are generated via rejection sampling: Draw selection vectors $s_i \sim \mathcal{U}$ i.i.d., compute $\mu(\mathbf{X}^{s_i})$, and accept $s_i$ if $\mu(\mathbf{X}^{s_i}) \le W$. Let $\{\overline{s}^1, \dots, \overline{s}^n\}$ be the first $n$ accepted selection vectors, yielding the masked inputs $\overline{\mathbf{X}}^c = \mathbf{X}^{\overline{s}^c}$. These samples $\{\overline{\mathbf{X}}^1, \dots, \overline{\mathbf{X}}^n\}$ are i.i.d.\ draws from the conditional distribution $\mathcal{D}_W$, defined as the distribution of $\mathbf{X}^s$ given $s \sim \mathcal{U}$ and $\mu(\mathbf{X}^s) \le W$.

For each accepted sample $c$, define $\mu_c = \mu(\overline{\mathbf{X}}^c)$ and the inverse-error weight $\mu'_c = \tfrac{1}{\mu_c + 1}$.
Then the empirical average sequence
$$
\chi_n(\alpha)
\;=\;
\frac{1}{n} \sum_{c=1}^n \bigl[\mu'_c\,\overline{\mathbf{X}}^c(\alpha)\bigr]
$$
converges pointwise almost surely (with probability that tends to 1) as $n \to \infty$ to the limit, which is the true expectation under the conditional distribution $\mathcal{D}_W$:
$$
\chi(\alpha) \;=\; \lim_{n\to\infty}\chi_n(\alpha)
\;=\;
\mathbb{E}_{\mathbf{X}' \sim \mathcal{D}_W}\bigl[\mu'(\mathbf{X}')\,\mathbf{X}'(\alpha)\bigr].
$$
This limit can be further decomposed using the law of total expectation:
$$
\chi(\alpha)
\;=\;
\mathbb{E}_{\mathbf{X}' \sim \mathcal{D}_W}\bigl[\mu'(\mathbf{X}') \,\bigm|\;\alpha \text{ retained in } \mathbf{X}' \bigr]
\,\times\,
\mathbb{P}_{\mathbf{X}' \sim \mathcal{D}_W}\bigl(\alpha \text{ retained in } \mathbf{X}' \bigr)
\,\times\,
\mathbf{X}(\alpha),
$$
where~\begin{itemize}
    \item $\alpha \text{ retained in } \mathbf{X}'$ means the chunk containing coordinate $\alpha$ is unmasked in $\mathbf{X}'$ (i.e., $\mathbf{X}'(\alpha) = \mathbf{X}(\alpha)$),
    \item $\mathbb{E}_{\mathbf{X}' \sim \mathcal{D}_W}[\cdot]$ denotes expectation over samples drawn from the conditional distribution $\mathcal{D}_W$,
    \item $\mathbb{P}_{\mathbf{X}' \sim \mathcal{D}_W}(\cdot)$ denotes probability under the conditional distribution $\mathcal{D}_W$.
\end{itemize}

\noindent\emph{Interpretation.}
Because $\mu(\mathbf{X}_{\text{arg}})$ can represent \emph{any} user-defined, model-based error metric (e.g.\ cross-entropy, MSE, negative log-likelihood, perceptual loss), the chunk-based sampling and weighting procedure highlights input coordinates $\alpha$ that consistently contribute to model outputs deemed ``good'' (low $\mu$) according to that metric. For instance, in \emph{landmark detection}, if $\mu(\mathbf{X}_{\text{arg}})$ measures the error between predicted and ground-truth keypoint heatmaps, masked inputs $\mathbf{X}'$ with low $\mu$ are those preserving essential visual cues for locating the landmark. Conversely, inputs with high $\mu$ obscure or remove those cues. The \mupax\ method thus naturally extends to \textbf{any dimension} and \textbf{any differentiable or non-differentiable error function}, serving as a unified explanation technique.

Intuitively, the limit $\chi(\alpha)$ quantifies the importance of coordinate $\alpha$ by combining three factors, conditioned on the subset of ``good'' selection vectors (those with $\mu(\mathbf{X}^s) \le W$):
\begin{enumerate}
    \item The original feature value $\mathbf{X}(\alpha)$.
    \item The probability $\mathbb{P}_{\mathbf{X}' \sim \mathcal{D}_W}(\alpha \text{ retained in } \mathbf{X}')$: How often is coordinate $\alpha$ retained within this set of good-performing masked inputs?
    \item The average ``goodness'' $\mathbb{E}_{\mathbf{X}' \sim \mathcal{D}_W}[\mu'(\mathbf{X}') \mid \alpha \text{ retained in } \mathbf{X}']$: When $\alpha$ is retained among these good inputs, how good (low error / high $\mu'$) is the model's output on average?
\end{enumerate}
Coordinates that are frequently retained in low-error inputs receive a higher importance score $\chi(\alpha)$.

A formal proof is provided in \textbf{Appendix A}.

\subsection{Computational Complexity Comparison}
\label{sec:complexity_comparison}

Let $C_f$ be the cost of a model forward pass, $C_{\nabla_f}$ the cost of a backward pass, and $\N{X}$ the number of samples/steps for method $X$.

\begin{itemize}
    \item \textbf{\mupax:} Cost is $\approx \mathbf{O(\N{\mupax} \times C_f)}$. Requires $\N{\mupax}$ inferences on perturbed inputs. $\N{\mupax}$ depends on the desired number of explanations $n$ and the acceptance probability $p_W$ ($\N{\mupax} \approx n/p_W$), tunable via threshold $W$. Crucially, the $\N{\mupax}$ inferences are \textbf{extremely parallel}, allowing significant wall-clock time reduction on parallel hardware (time $\propto \N{\mupax}/P$ with $P$ processors). Sorting, if required, involves $O(N \log N)$ complexity.
    \item \textbf{LIME / KernelSHAP:} Cost $\approx \mathbf{O(\N{LIME} \times C_f)}$ / $\mathbf{O(\N{SHAP} \times C_f)}$. Also perturbation-based and parallelizable over samples $\N{LIME} \, / \, \N{SHAP}$. The relative cost vs.\ \mupax depends on the required sample sizes ($\N{\mupax}$ vs.\ $\N{LIME}\, / \, \N{SHAP}$) for comparable explanation quality.
    \item \textbf{Integrated Gradients (IG):} Cost $\approx \mathbf{O(\N{IG} \times C_{\nabla_f})}$. The sequential gradient computation is required $\N{IG}$  (typically $\N{IG} \ll \N{\mupax}$). It is less parallelizable across the $\N{IG}$ steps.
    \item \textbf{Grad-CAM:} Cost $\approx \mathbf{O(C_f + C_{\nabla_f})}$. Very fast, requiring only one forward and one backward pass. Less general (requires specific architectures).
\end{itemize}

 \textbf{Comparative View:} While gradient methods (Grad-CAM, IG) are often faster in total FLOPS, \mupax\ offers model-agnosticism and a direct simulation approach to identify sufficient inputs. Its primary cost, $\N{\mupax}$ inferences, is highly parallelizable. This means its practical runtime can be competitive with, or even faster than, other perturbation methods (LIME, KernelSHAP) on modern hardware, especially if high fidelity requires large sample counts ($\N{LIME}$, $\N{SHAP}$) for those methods as well. Furthermore, the cost $\N{\mupax}$ is tunable via the threshold $W$, providing a direct mechanism to trade computational budget for explanation granularity. This parallel nature and tunability make \mupax\  potentially appealing despite the potentially large total number of operations.
 There is a trade-off between explanation fidelity and computational cost. The total number of forward passes, $\N{\mupax} = n / p_W$, depends on desired samples $n$ and acceptance probability $p_W$, regulated by threshold $W$. Lowering $W$ increases fidelity (more discriminative sample selection) but lowers $p_W$, and hence raises $\N{\mupax}$. Users have to balance $W$ and $n$ according to application needs and available resources, accepting potential differences in precision for efficiency.

\section{Experiments and Results}
\label{sec:experiments}

\subsection{Dataset and Model Training}

To show the versatility of the \mupax\ framework, we vary the data modality, the dimensionality of the data, as well as the application tasks. For each dataset below, a model underwent training, followed by freezing upon completion. Subsequently, explainability methods appropriate for the data and task were evaluated and compared to the \mupax\ framework. All the experiments were executed on a PC with AMD Ryzen™ Threadripper™ 3970X × 64 with 128GB of DDR4 RAM, \textbf{a single Nvidia A6000 GPU with 48GB DDR6 VRAM} and with Ubuntu 24.04.2 LTS. 

We used four publicly available datasets:

\begin{itemize}
    \item \textbf{1D Audio Signal Multi-class Classification:} We used the GTZAN dataset \cite{tzanetakis2002gtzan}, which comprises $1,000$ audio tracks categorized into $10$ genres. Each audio file had a duration of $3$ seconds and was recorded at 16-bit resolution with a sampling rate of $22,050$ Hz. We transformed the audio signals into RGB Mel-spectrogram images of resolution of $775\times308$ pixels. Subsequently, we trained a ResNet-50 model using these spectrogram images, applying a window size of $2048$, a hop length of $344$, and employing $150$ Mel frequency bins. We adapted a ResNet-50 model, pretrained on ImageNet, by replacing its final fully connected layer with a softmax classifier over the $10$ genre classes. Optimization during training was performed using the RMSProp algorithm, with data partitioned into a $90:10$ train-test split.
    
    \item \textbf{2D Binary Image Classification:} We used the Cats vs.\ Dogs dataset \cite{catvsdog}, which contains 3,000 images equally divided between the two classes. The original images varied considerably in resolution, with a mean size of $410.2 \times 364.9$ pixels. For consistency in training, we resized all inputs to $128\times128$ pixels. We employed a ConvNeXtXLarge backbone pretrained on ImageNet, followed by a global average pooling layer, a fully connected layer with $1,024$ ReLU-activated units, and a final softmax output layer. The model was trained end-to-end on the dataset, which we split into training and test sets in an $80:20$ ratio.
    
    \item \textbf{3D Volume Binary Classification:} We employed the MosMed CT dataset \cite{morozov2020mosmeddata}, which comprises $1,110$ pulmonary computed tomography (CT) volumes used to distinguish COVID-19-positive cases from non-COVID cases. Each volume consists of axial slices with a resolution of $512\times512$ pixels and a slice thickness of $8.00$ mm, with the number of slices varying between $33$ and $52$ per volume. We developed a custom 3D convolutional neural network comprising three Conv3D blocks with $64$, $128$, and $256$ filters, respectively, each followed by max pooling and batch normalization layers. These were followed by a global average pooling layer, a dense layer with $512$ ReLU-activated units, a dropout layer, and a final softmax output layer. The dataset was split $70:30$.

    \item \textbf{2D Image Landmark Detection:} We used the CephAdoAdu dataset \cite{wu2024cephalometric}, which includes $1,000$ cephalometric X-ray images evenly divided between $500$ adult and $500$ adolescent patients. Each image is annotated with $10$ clinically significant anatomical landmarks, identified by an expert dental radiologist. The original images vary considerably in resolution, with a mean of $1,929 \times 1,853$ pixels. For the training, we rescaled all images to a standardized resolution of $256 \times 256$ pixels. We employed a ConvNeXt Tiny backbone \cite{liu2022convnet} pretrained on ImageNet, integrated with a decoder composed of four upsampling layers with $256$, $128$, $64$, and $32$ filters, respectively. Each upsampling step was followed by convolutional blocks, and a final $1\times1$ convolutional layer generated heatmaps for each landmark, which were subsequently resized to the target resolution via bilinear interpolation. The dataset was randomly split $80:20$, with equal representation of adult and adolescent cases to ensure balanced age group coverage.

\end{itemize}

\begin{figure}[t]
\centering

    \begin{subfigure}{\imsize\textwidth}
    \centering
    \includegraphics[width=\textwidth]{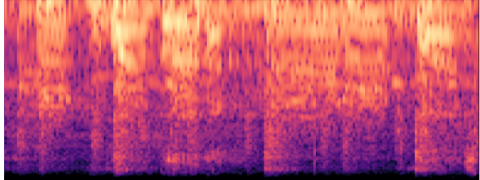}
    \caption{\footnotesize{Orig.\\Pred: `blues'}}
    \end{subfigure}
    \hfill
    \begin{subfigure}{\imsize\textwidth}
    \includegraphics[width=\textwidth]{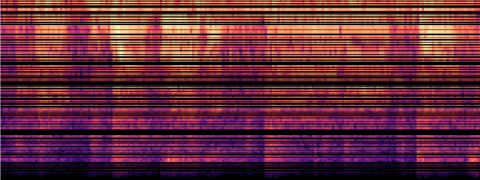}
    \caption{\footnotesize{\mupax\ \\Pred: `blues'}}
    \end{subfigure}
    \begin{subfigure}{\imsize\textwidth}
    \includegraphics[width=\textwidth]{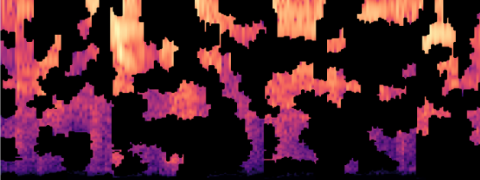}
    \caption{\footnotesize{LIME \\Pred: `hiphop'}}
    \end{subfigure}
    \begin{subfigure}{\imsize\textwidth}
    \includegraphics[width=\textwidth]{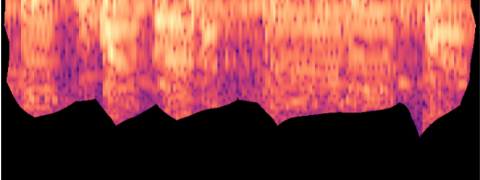}
    \caption{\footnotesize{GradCAM \\Pred: `blues'}}
    \end{subfigure}
    \begin{subfigure}{\imsize\textwidth}
    \includegraphics[width=\textwidth]{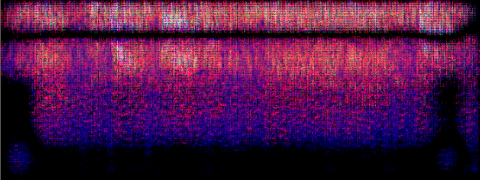}
    \caption{\footnotesize{SHAP \\Pred: `classical'}}
    \end{subfigure}
    \begin{subfigure}{\imsize\textwidth}
    \includegraphics[width=\textwidth]{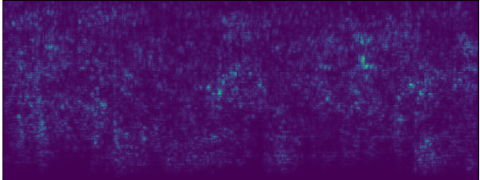}
    \caption{\footnotesize{Int. grads.\\Pred: `Jazz'}}
    \end{subfigure}

    \caption{1D audio classification explanation: \mupax\ outperforms LIME and GradCAM in identifying important frequency features.}
    \label{fig:1D_audio}

\end{figure}

\begin{figure}[t]
\centering

    \begin{subfigure}{\imsize\textwidth}
    \centering
    \roundedimage{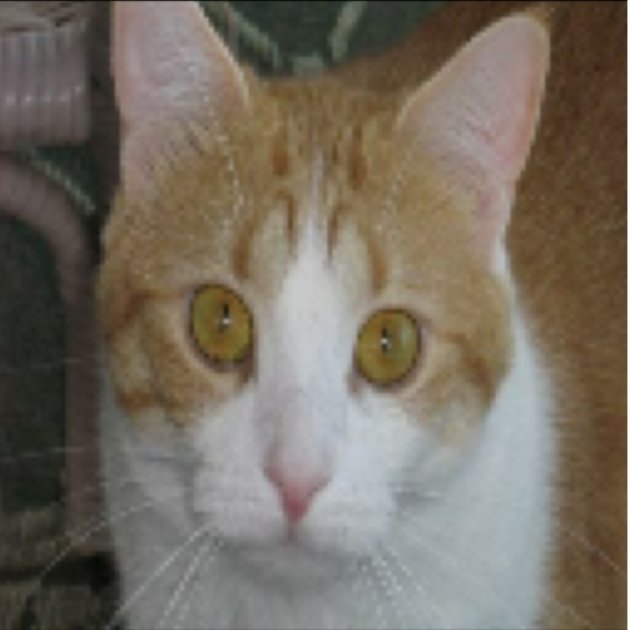}{\textwidth}
    \caption{\footnotesize{Orig.\\Pred: `cat'}}
    \end{subfigure}
    \hfill
    \begin{subfigure}{\imsize\textwidth}
    \roundedimage{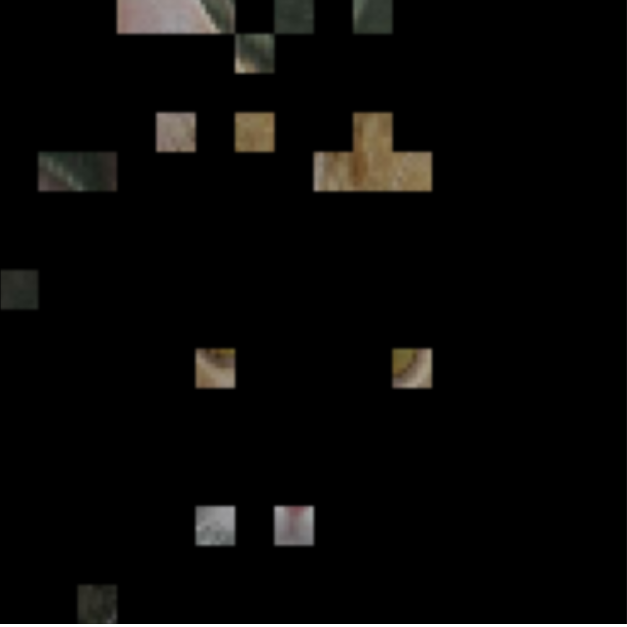}{\textwidth}
    \caption{\footnotesize{\mupax\\Pred: `cat'}}
    \end{subfigure}
    \begin{subfigure}{\imsize\textwidth}
    \roundedimage{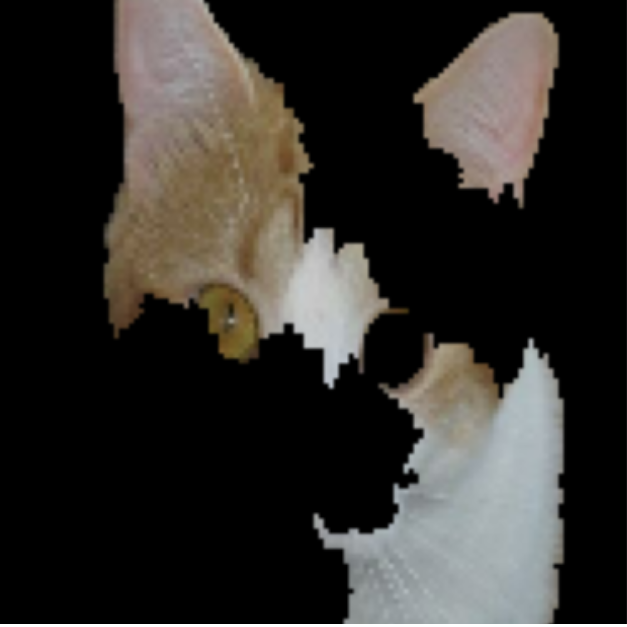}{\textwidth}
    \caption{\footnotesize{LIME\\Pred: `cat'}}
    \end{subfigure}
    \begin{subfigure}{\imsize\textwidth}
    \roundedimage{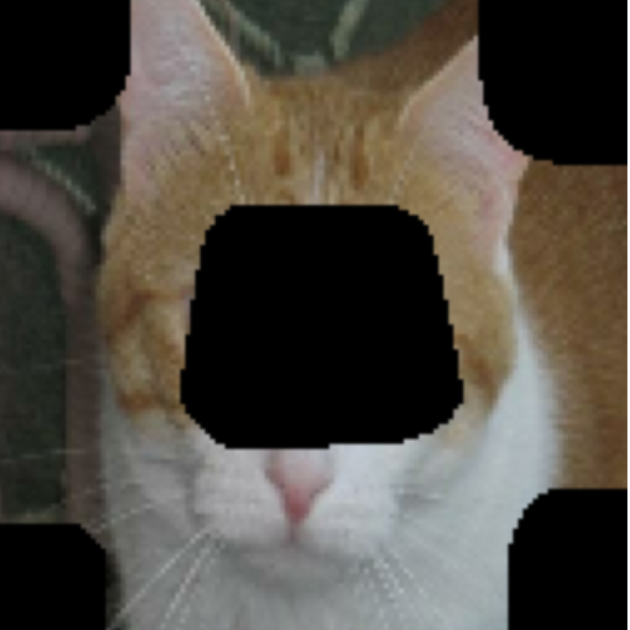}{\textwidth}
    \caption{\footnotesize{GradCAM\\Pred: `dog'}}
    \end{subfigure}
    \begin{subfigure}{\imsize\textwidth}
    \roundedimage{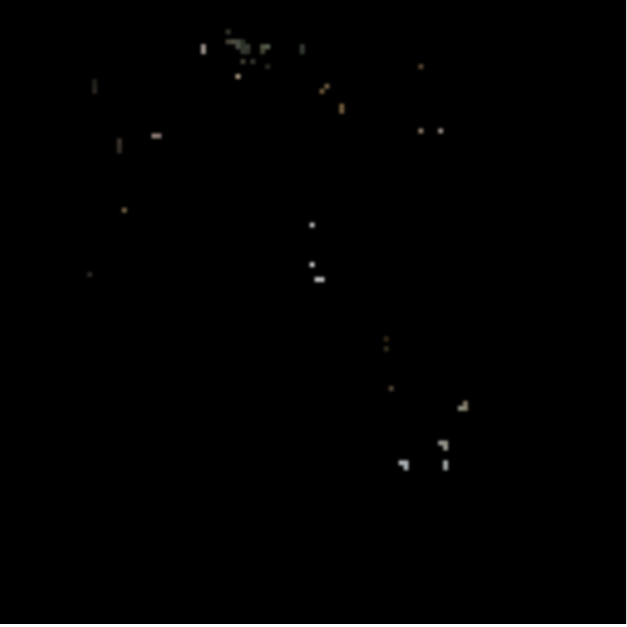}{\textwidth}
    \caption{\footnotesize{SHAP\\Pred: `dog'}}
    \end{subfigure}
    \begin{subfigure}{\imsize\textwidth}
    \roundedimage{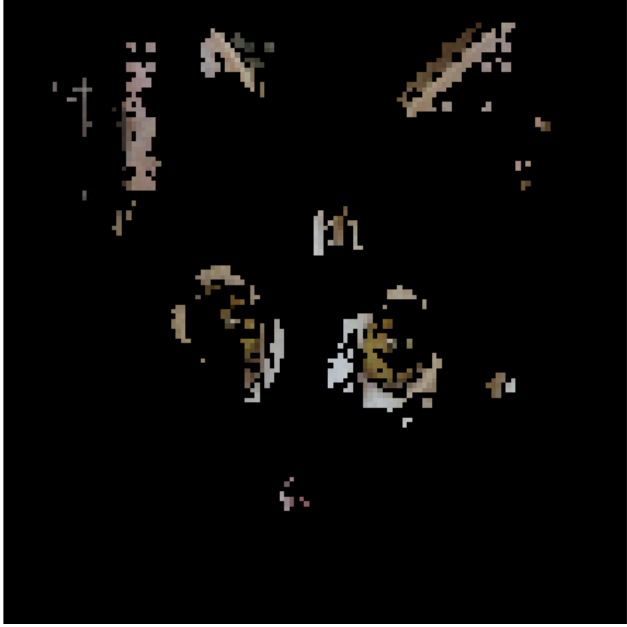}{\textwidth}
    \caption{\footnotesize{Int. grads.\\Pred: `dog'}}
    \end{subfigure}

    \caption{2D cat-dog image classification: \mupax\ preserves discriminative patterns w.r.t other methods.}
    \label{fig:2D_image}

\end{figure}

\subsection{Classification Methodology for Filter Mask Evaluation}

Filter mask experiments evaluated the performance of existing XAI methods to identify significant features in a variety of contexts. A separate model was trained and frozen for each database. XAI methods LIME, GradCAM, SHAP, and Integrated Gradients  generated attribution maps that expose salient regions for classification, which were compared against \mupax.

For the 1D audio signal classification, illustrated in Figure~\ref{fig:1D_audio}, \mupax\ was executed only on the frequency axis of the mel spectrogram with 500 perturbation samples, whereas LIME, GradCAM, and SHAP were executed with their default implementations and default threshold values at the $50^{th}$ percentile to highlight significant regions.

For 2D image classification, illustrated in Figure~\ref{fig:2D_image}, \mupax\ segmented images into $8\times8$ pixel patches and took 2000 perturbation samples. LIME produced 100 superpixels with 500 perturbation samples. GradCAM utilized the final convolutional layer of ConvNeXtXLarge and took $50^{th}$ percentile threshold attributions. SHAP employed a Gradient Explainer with 100 background samples, and Integrated Gradients utilized 50 steps of interpolation and a baseline of zero.

\begin{figure}[t]
\centering

    \begin{subfigure}{\imsize\textwidth}
    \centering
    \includegraphics[width=\textwidth]{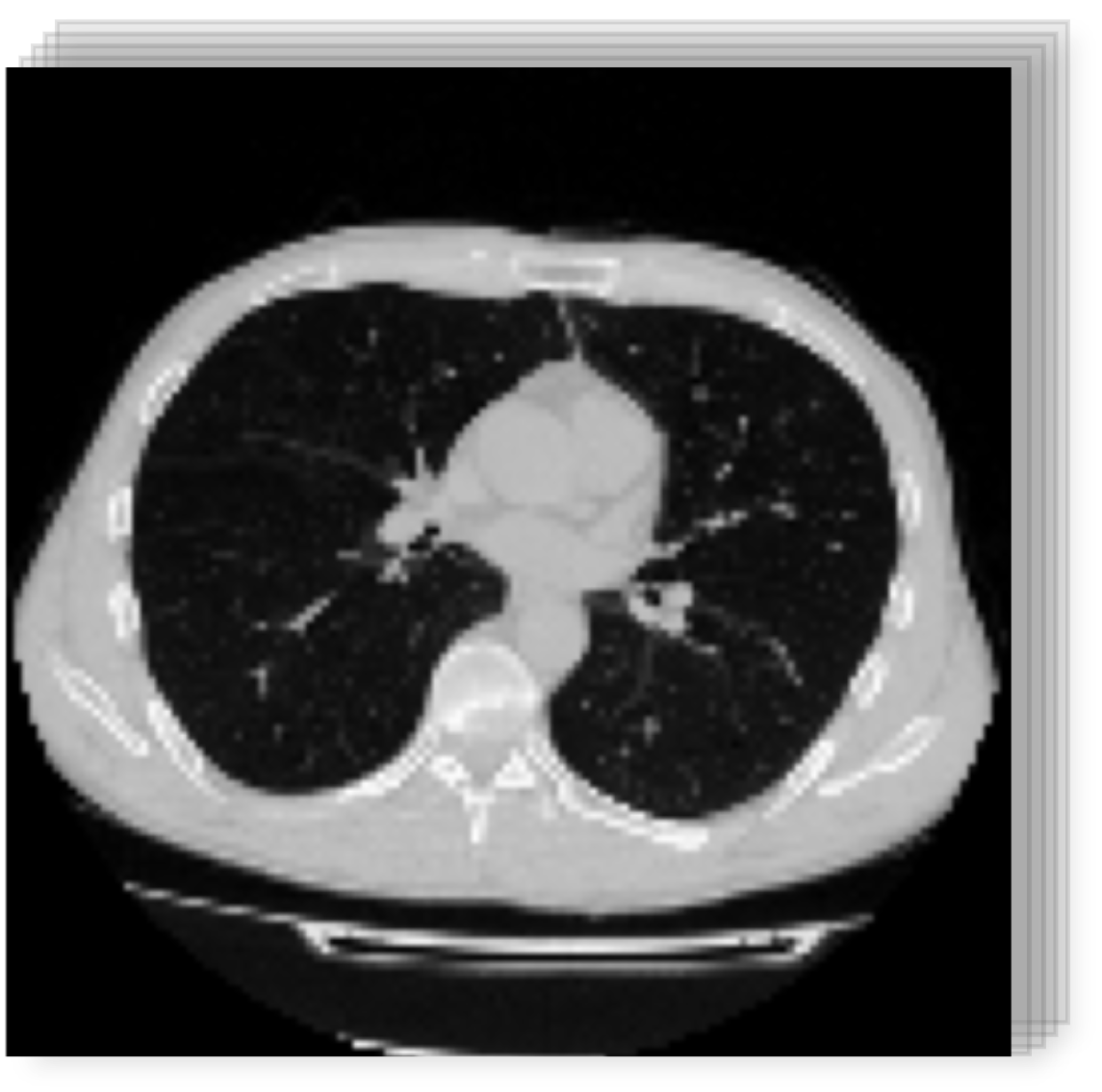}
    \caption{\footnotesize{Orig.\\Pred: `Normal'}}
    \end{subfigure}
    \hspace{5pt}
    \begin{subfigure}{\imsize\textwidth}
    \includegraphics[width=\textwidth]{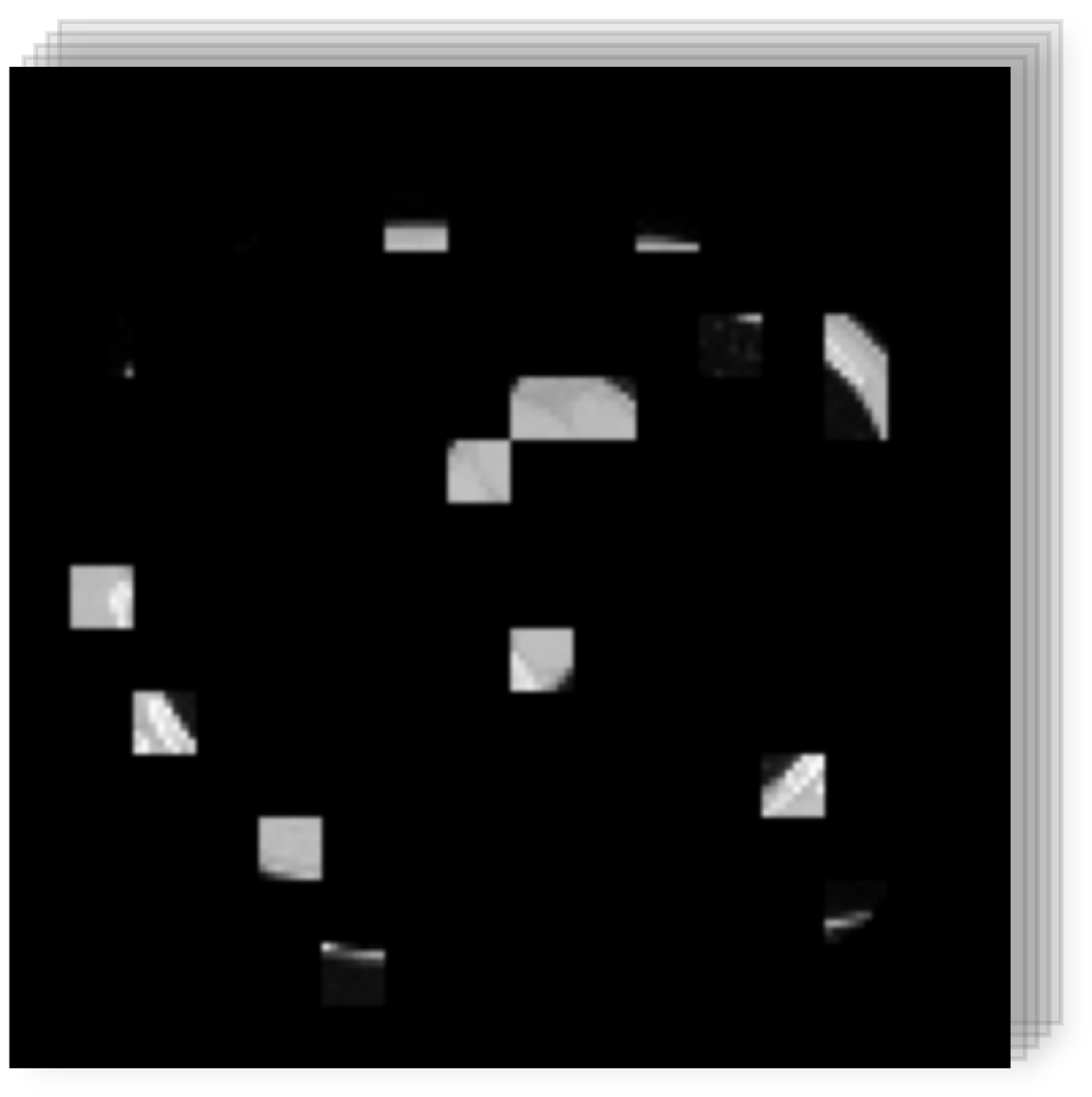}
    \caption{\footnotesize{\mupax\\Pred: `Normal'}}
    \end{subfigure}
    \begin{subfigure}{\imsize\textwidth}
    \includegraphics[width=\textwidth]{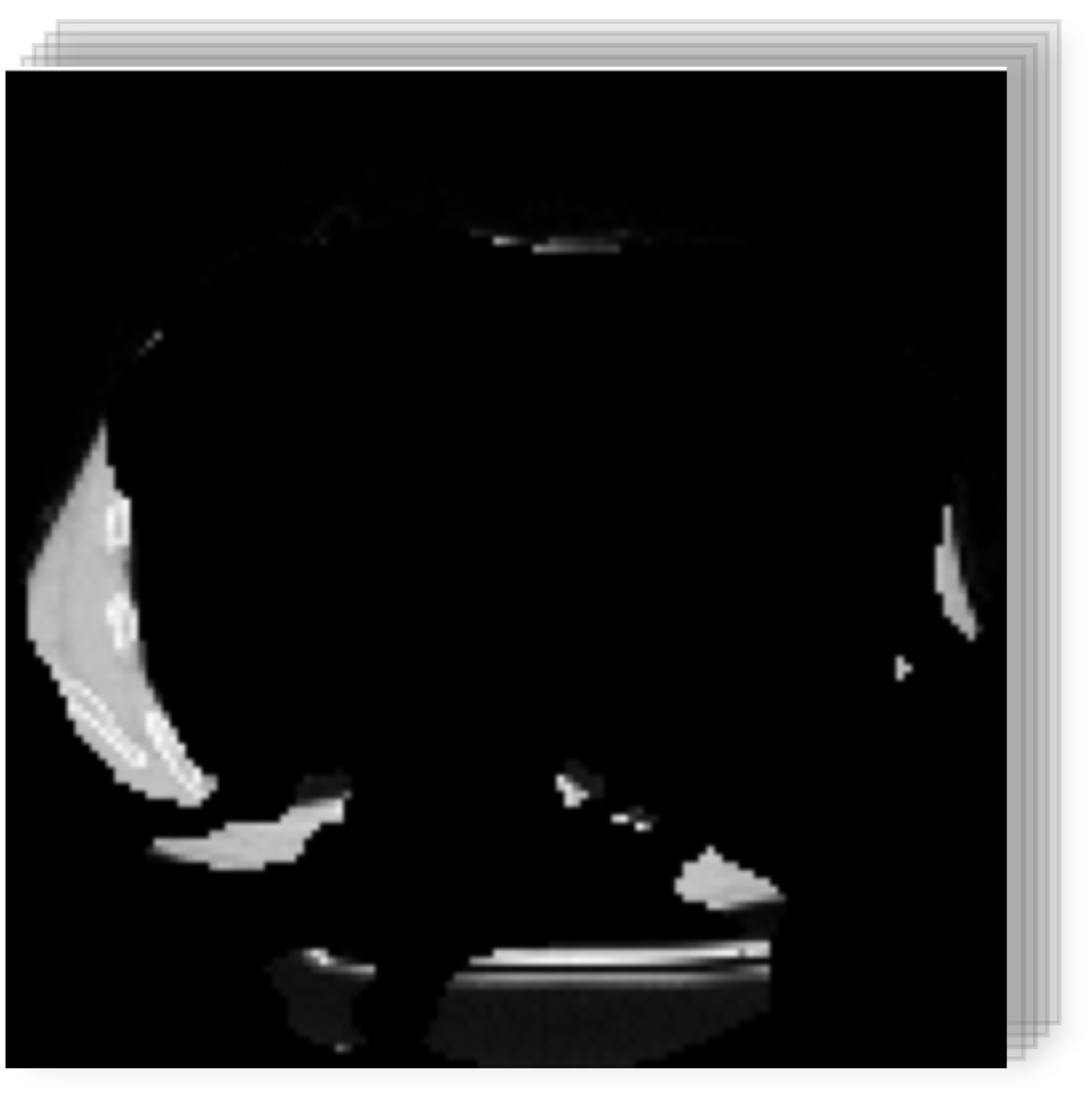}
    \caption{\footnotesize{GradCAM\\Pred: `Abnormal'}}
    \end{subfigure}
    \begin{subfigure}{\imsize\textwidth}
    \includegraphics[width=\textwidth]{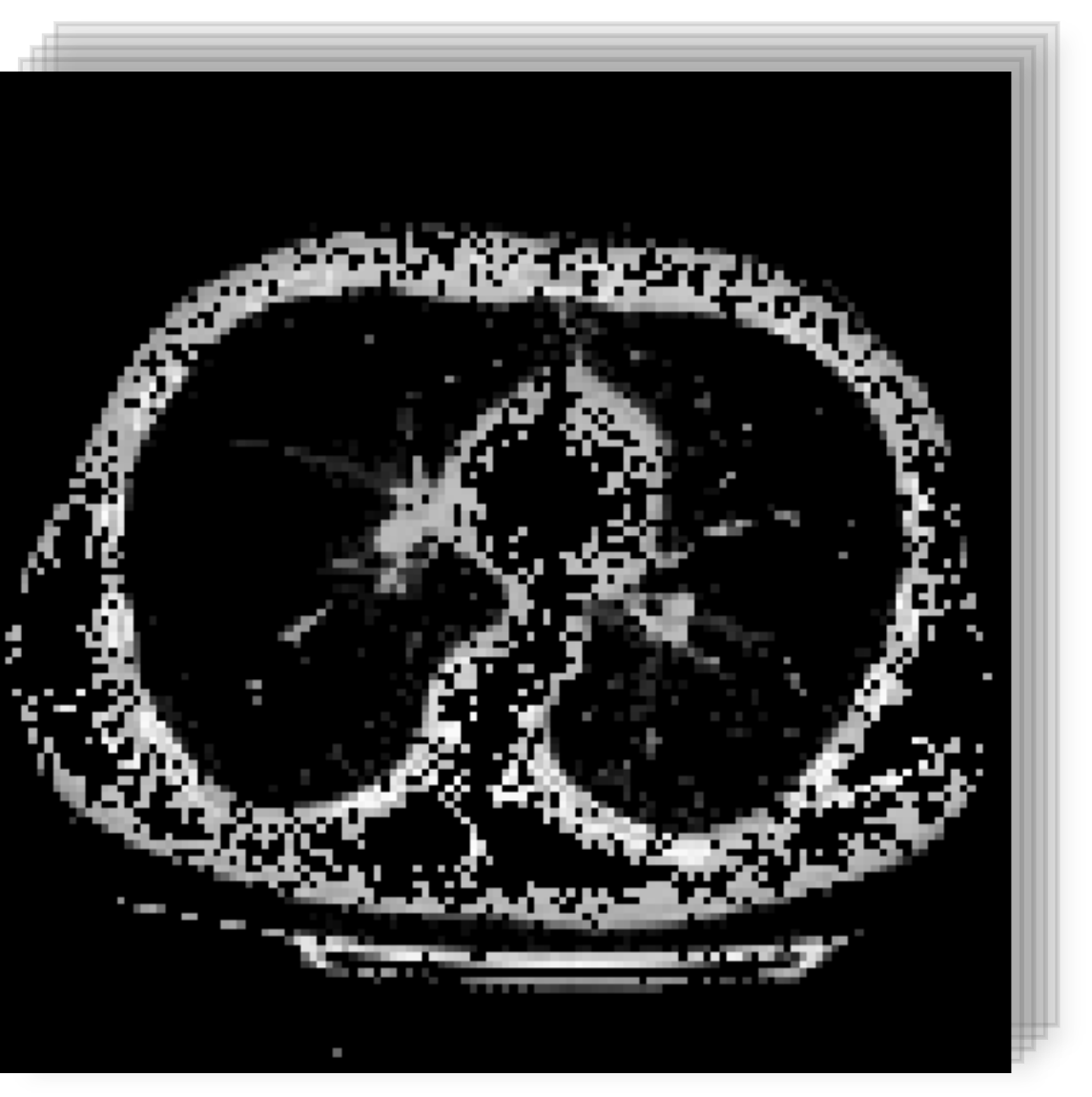}
    \caption{\footnotesize{Int. grads.\\Pred: `Abnormal'}}
    \end{subfigure}
    \caption{3D COVID-19 CT classification explanation: \mupax\ highlights diagnostic features (F1=0.88) vs.\ GradCAM (0.42) and Integrated Gradients (0.33).}
    \label{fig:3D_ct}

\end{figure}

For 3D volume classification, illustrated in Figure~\ref{fig:3D_ct}, \mupax\ partitioned CT volumes into $8\times8\times8$ patches with 200 perturbation samples. GradCAM used the last convolutional layer of the 3D CNN with a $50^{th}$ percentile threshold, and Integrated Gradients computed attributions over 50 steps with a zero baseline. LIME and SHAP cannot deal with 3D volumes.

To quantify effectiveness, regions that were regarded as irrelevant by all XAI techniques were masked out, to produce filtered inputs. Models were then tested on the filtered inputs to quantify the change in classification performance when using only features deemed relevant. Classification performance metrics (precision, recall, macro and weighted F1-scores) were calculated for the classification output of every method.
Experimental results are summarised in Table~\ref{tab:combined_classification_results}.

\mupax\ remains superior to the other state-of-the-art methods across the board, the only method able not only to maintain, but also to improve classification performance by removing irrelevant features with respect to the baseline, i.e. the original frozen model.
On the 1D audio classification task, \mupax\ maintained classification integrity with a macro F1-score of 0.51, which is similar to the baseline, and far superior to LIME (0.16) and GradCAM (0.29). This suggests that \mupax\ chooses  frequency elements which are unequivocally  significant to genre classification, while other approaches over-filter useful information or retain irrelevant traits.

The contrast is all the more pronounced in the case of 2D image classification, since \mupax\ 
improves the baseline F1-score from 0.93 to 0.95 with masking. It appears that \mupax\ is effective in noise removal and potentially distracting visual data, allowing the model to focus only on discriminating features. GradCAM (0.47), SHAP (0.45), and Integrated Gradients~(0.66), for comparison, all result in a dramatic performance plunge in all metrics relative to the baseline.

For 3D medical image volumes, the trend is similar, with \mupax\ improving classification the F1-score from 0.82 to 0.88, while GradCAM (0.42) and Integrated Gradients (0.33) severely declined in performance relative to the baseline.
This carries particular significance in medical use, where feature accuracy translates  into diagnostic consistency and clinician trust.

Table~\ref{tab:combined_classification_results} also quotes, in seconds, the time taken to run each method.

To complement the mask-based evaluation,  faithfulness has also been assessed by measuring the impact of perturbing the \emph{most important\/} features identified by each XAI method. Following standard practice, it is possible to identify the top  features according to each method's attribution map and replace their values (e.g.\ with zero or a baseline value), then measure the degradation in model performance (e.g.\ drop in prediction probability for the original class).

\begin{table}[t]
    \centering
    \footnotesize
    \caption{Comparison of XAI Classification Performance Across Different Data Dimensions}
    \label{tab:combined_classification_results}
    \begin{tabular}{ccccccc}
        \toprule
            \textbf{Dimension} & 
            \textbf{Method} & 
            \textbf{Precision} & 
            \textbf{Recall} & 
            \textbf{Macro F1} & 
            \textbf{Weighted F1} & 
            \textbf{\begin{tabular}[c]{@{}c@{}}Runtime \\(sec)\end{tabular}}\\
        \midrule
        \multirow{5}{*}{\textbf{1D Signal}} 
        & Full input & \textbf{0.52} & 0.47 & 0.45 & 0.45 \\
        & \mupax mask & 0.51 & \textbf{0.61} & \textbf{0.51} & \textbf{0.51} & 06.66 ± 0.68 \\
        & LIME mask & 0.22 & 0.22 & 0.16 & 0.16 & 14.73 ± 0.55 \\
        & GradCAM mask & 0.48 & 0.36 & 0.29 & 0.29 & 00.19 ± 0.02 \\
        & SHAP mask & 0.49 & 0.50 & 0.47 & 0.47 & 29.86 ± 1.32 \\
        & Integrated gradients & 0.50 & 0.44 & 0.41 & 0.41 & 07.10 ± 0.13 \\
        \midrule
        \multirow{6}{*}{\textbf{2D Image}} 
        & Full Image & 0.93 & 0.93 & 0.93 & 0.93 \\
        & \mupax mask & \textbf{0.95} & \textbf{0.95} & \textbf{0.95} & \textbf{0.95} & 06.27 ± 1.57 \\
        & LIME mask & 0.90 & 0.91 & 0.90 & 0.90 & 04.40 ± 0.97 \\
        & GradCAM mask & 0.47 & 0.47 & 0.47 & 0.48 & 05.66 ± 0.33 \\
        & SHAP mask & 0.53 & 0.58 & 0.45 & 0.62 & 29.46 ± 1.67 \\
        & Integrated gradients & 0.76 & 0.69 & 0.66 & 0.66 & 07.37 ± 0.84 \\
        \midrule
        \multirow{4}{*}{\textbf{3D Image}} 
        & Full Image & 0.82 & 0.82 & 0.82 & 0.82 \\
        & \mupax mask & \textbf{0.91} & \textbf{0.88} & \textbf{0.88} & \textbf{0.88} & 09.16 ± 0.45 \\
        & GradCAM mask & 0.50 & 0.50 & 0.42 & 0.42 & 00.23 ± 0.31 \\
        & Integrated gradients & 0.25 & 0.50 & 0.33 & 0.33 & 02.36 ± 0.82 \\
        \bottomrule
    \end{tabular}
\end{table}

\subsection{Landmark detection}
\label{subsec:landmark_detection}

\mupax\ comes into its own when applied to structured‑output tasks such as landmark detection. 
\mupax\ for landmark detection partitioned the images into $16\times16$ patches with 500 perturbation samples.  GradCAM, as before, used the last convolution layer. 
For each test image and for each individual landmark, approximately 41–42\% of the original pixels  are retained into the \mupax‑based crop. The landmark predictions obtained via \mupax\ are compared to predictions using full images and GradCAM-cropped images.
Table \ref{tab:landmark_errors} reports mean radial error. Due to the nature of the dataset (no fixed millimetre per pixel ratio) all measurements are reported in pixels (px).
Radial error is listed for three cases: using the full original image, the \mupax\-cropped image and GradCAM-cropped images. From the related works listed in Section~\ref{sec:related}, only Class Activation Map methods are applicable to landmark detection. Even these, having never been designed for this task, struggle to contain the relevant search area, but GradCAM was included in these experiments for completeness. 

The radial error for nine out of ten landmarks is lowered when using a \mupax-based landmark detection model~\cite{liu2022convnet}, reducing the overall average from $\sim$13.8 px to $\sim$9.3 px. The largest improvements occur at landmarks 6 and 8 (improvements of 13.8 px and 12.0 px, respectively), corresponding to anatomically compact regions successfully isolated by chunk‑based selection. A small error increase is observed for landmark 9, located on the nose, where partial omission of peripheral cues can occur.

\begin{figure}[t]
\centering

    \begin{subfigure}{\imsize\textwidth}
    \centering
    \includegraphics[width=\textwidth]{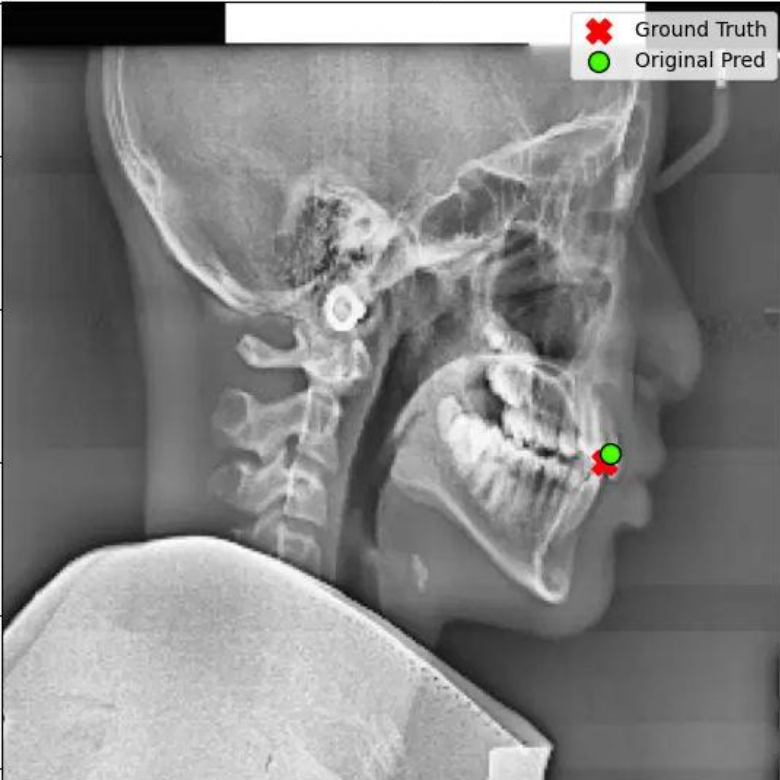}
    \caption{\footnotesize{Orig.\\Pred}}
    \end{subfigure}
    \begin{subfigure}{\imsize\textwidth}
    \includegraphics[width=\textwidth]{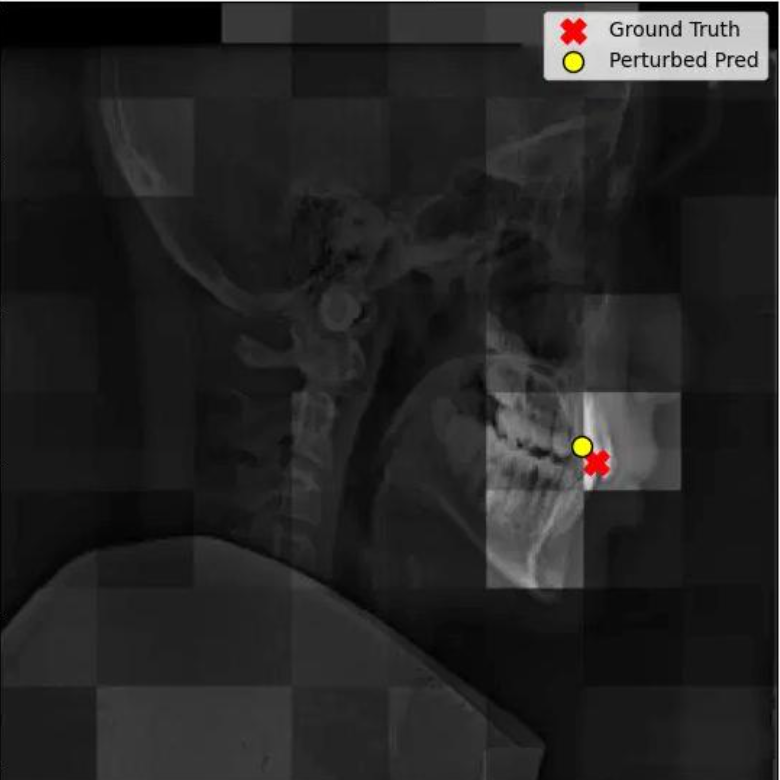}
    \caption{\footnotesize{\mupax\\Pred}}
    \end{subfigure}
    \begin{subfigure}{\imsize\textwidth}
    \includegraphics[width=\textwidth]{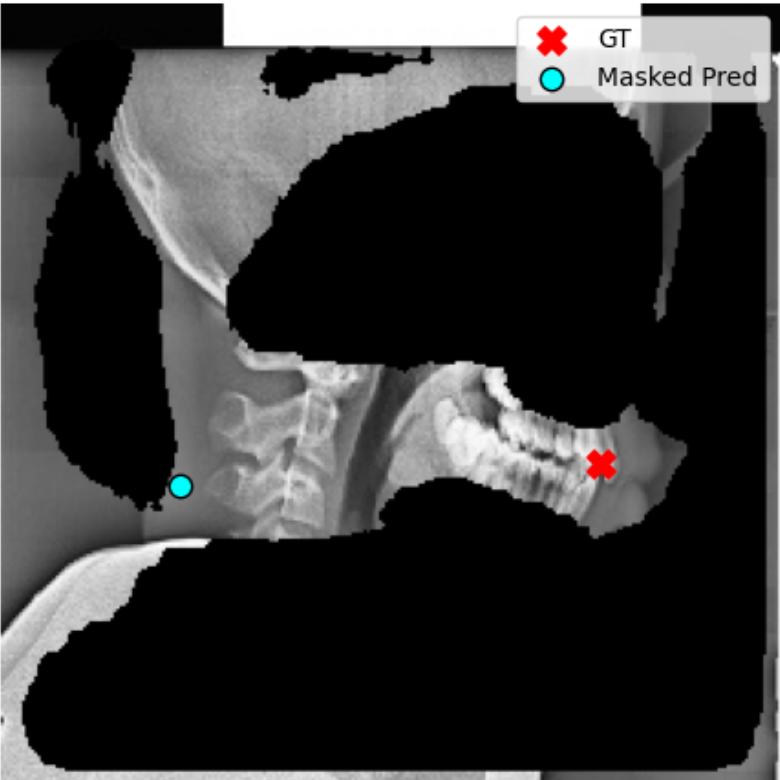}
    \caption{\footnotesize{GradCAM\\Pred}}
    \end{subfigure}
    \hspace{5pt}
    \begin{subfigure}{\imsize\textwidth}
    \centering
    \includegraphics[width=\textwidth]{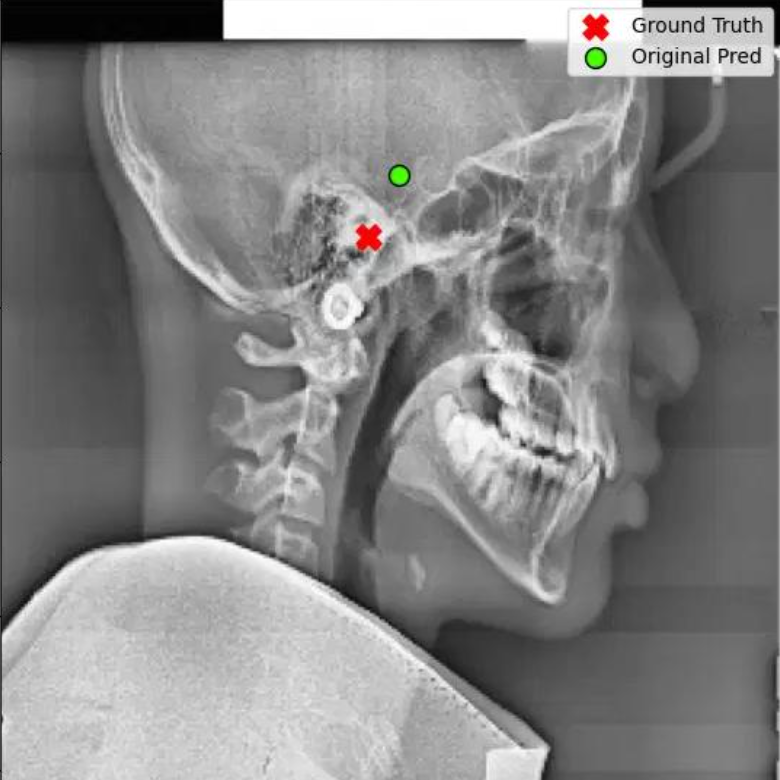}
    \caption{\footnotesize{Orig.\\Pred}}
    \end{subfigure}
    \begin{subfigure}{\imsize\textwidth}
    \includegraphics[width=\textwidth]{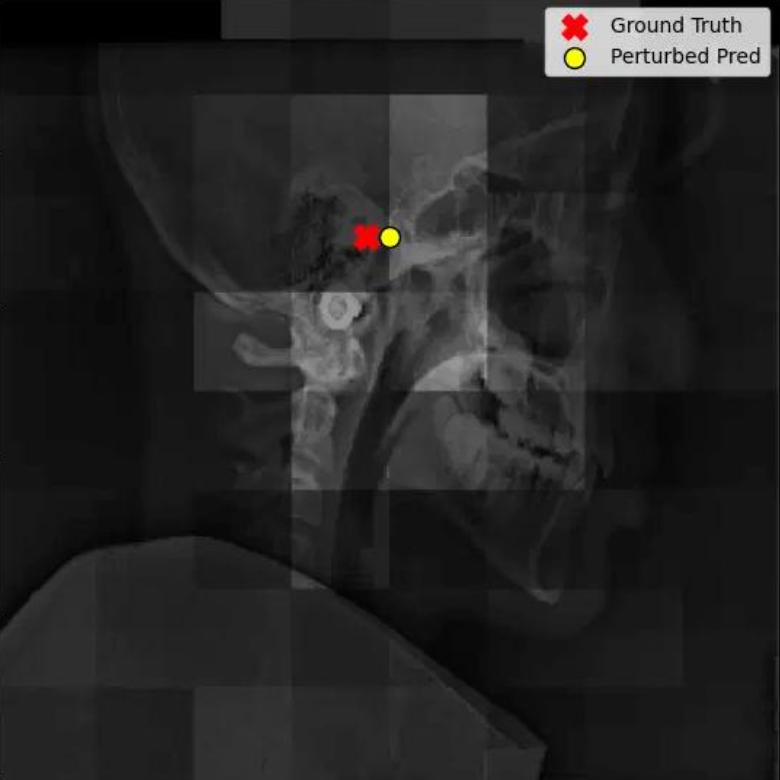}
    \caption{\footnotesize{\mupax\\Pred}}
    \end{subfigure}
    \begin{subfigure}{\imsize\textwidth}
    \includegraphics[width=\textwidth]{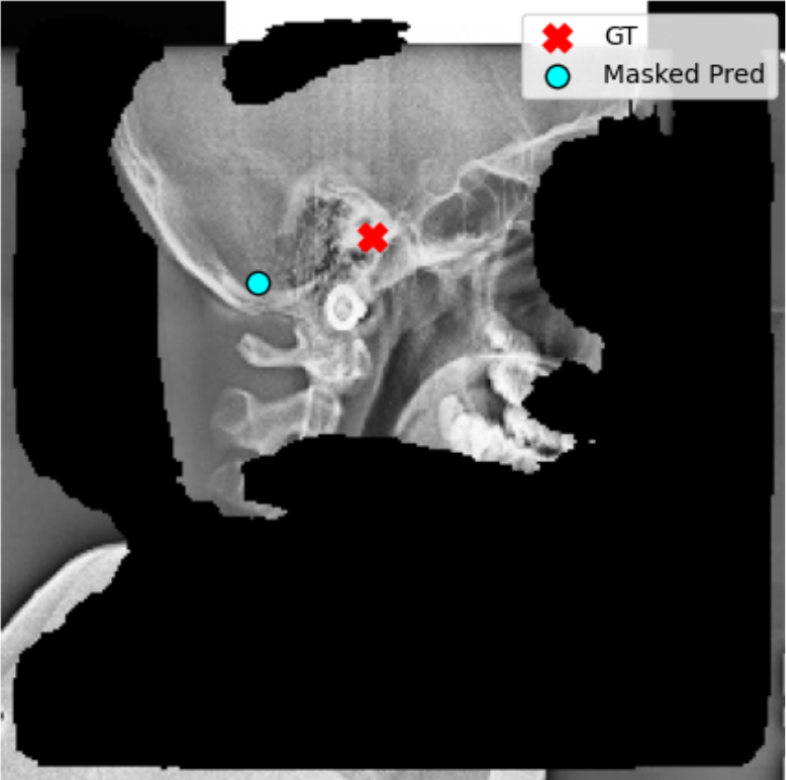}
    \caption{\footnotesize{GradCAM\\Pred}}
    \end{subfigure}
    
    \caption{Landmark detection qualitative output for Landmarks 7 (left) and 6 (right) from using the full original image, together with areas highlighted in the use of \mupax\ and GradCAM. Red cross: the human-annotated landmark, in green: the full image prediction, in yellow the \mupax\-cropped image prediction, in cyan the GradCAM-cropped image prediction (all extended to 6 px radius).}
    \label{fig:combined_cephalometric}

\end{figure}

\begin{table}[t]
    \centering
    \footnotesize
    \caption{Landmark detection radial error (mean $\pm$ standard deviation) using full original images vs.\ images cropped with \mupax\ or GradCAM.}
    \label{tab:landmark_errors}
    \begin{tabular}{lccccc}
        \toprule
        \multirow{2}{*}{
        \textbf{Landmark}
        } & 
        \textbf{
        Orig.
        } & 
        \multicolumn{2}{c}{\textbf{\mupax}} & 
        \multicolumn{2}{c}{\textbf{GradCAM}} \\
         &
        Error (px) &
        Error (px) &
        \% used &
        Error (px) &
        \% used 
        \\
\midrule
        ~~1. Subspinale  & 05.60 ± 02.49 & \textbf{04.83 ± 02.74} & 41.41\% & 113.25 ± 082.06 & 40.0\% \\
        ~~2. Anterior nasal spine  & 05.46 ± 02.58 & \textbf{04.75 ± 02.56} & 41.80\% & 121.18 ± 056.69 & 40.0\% \\
        ~~3. Incision superus & 09.38 ± 03.43 & \textbf{05.95 ± 08.24} & 42.03\% & 055.47 ± 047.59 & 40.0\% \\
        ~~4. Upper Incisor apex & 12.46 ± 02.84 & \textbf{07.14 ± 03.03} & 42.29\% & 026.18 ± 007.04 & 40.0\% \\
        ~~5. Orbitale & 18.25 ± 05.87 & \textbf{11.78 ± 06.43} & 42.50\% & 046.14 ± 017.37 & 40.0\% \\
        ~~6. Porion & 35.01 ± 10.42 & \textbf{21.22 ± 14.92} & 42.40\% & 035.13 ± 012.53 & 40.0\% \\
        ~~7. Incision inferius & 08.75 ± 14.02 & \textbf{07.53 ± 13.10} & 42.00\% & 075.91 ± 042.37 & 40.0\% \\
        ~~8. Lower Incisor apex & 24.30 ± 05.98 & \textbf{12.30 ± 09.60} & 42.24\% & 047.21 ± 038.97 & 40.0\% \\
        ~~9. Subnasale & \textbf{05.87 ± 02.76} & 06.12 ± 04.70 & 42.33\% & 115.77 ± 064.07 & 40.0\% \\
        10. Soft tissue pogonion & 12.86 ± 05.98 & \textbf{11.38 ± 09.53} & 40.56\% & 066.88 ± 044.70 & 40.0\% \\
        \midrule
        Overall & 13.79 ± 09.59 & 09.30 ± 05.07 & 41.96\% & 070.31 ± 035.06 & 40.0\% \\
        \bottomrule
    \end{tabular}
\end{table}

For clarity reasons, the runtimes are not included in Table~\ref{tab:landmark_errors}, but the complexity trend 
is unsurprising. \mupax\ runs in $4.44\pm0.29$s per image, versus $2.78\pm0.09$s for GradCAM, echoing the earlier results where \mupax\ consistently trades a modest runtime premium for markedly higher explanatory fidelity and accuracy.

Figure \ref{fig:combined_cephalometric} illustrates a qualitative evaluation of Landmarks 7 and 6, the best-- and worst--performing ones from the set, considering, a random image chosen from the set. Each is illustrated through three images: the original image, the \mupax\ prediction and the GradCAM prediction. The regions highlighted in the middle and right of each triplet represent the regions retained by \mupax\ and GradCAM respectively.

These results demonstrate that, even when cropping the search space down to less than half of the pixel area, the \mupax‑based crops still improve thelandmark localization. \mupax\ identifies and excludes regions containing noise or distracting features, retaining only those image parts that drive correct landmark predictions. Looking at the highlighted subjectively, it is credible, in each case, that the landmark was identifiable using that reduced set of pixels, without looking further afield. It is also unsurprising that the search area for Landmark~6 is larger. Even in human terms, since it is a place where the bone is more uniform, its precise locus is less precisely defined.

Ablation studies and discussion on varying core hyperparameters for all tested techniques can be found in \textbf{Appendix B}.

\section{Discussion}
\label{sec:discussion}

Gradient data is known to be prone to model architecture idiosyncrasies such as gradient saturation, architectural nuances, noisy backpropagation that can compromise gradient‐based explanations, etc.
Rather than being driven by gradient data, \mupax's perturbation-based approach methodically seeks feature importance throughout the entire input space. 
It inherently identifies and removes noisy or non-generalizable features (spurious correlations). Since the underlying model may have taken these into account when solving the task, their removal acts as a form of post-hoc regularization which, in turn, helps the model focus on more relevant patterns. Further investigation is needed to fully understand this effect, but to note is that a welcome side-effect of this removal is also to improve overall performance.
\mupax\ also achieves a trade-off between pinpoint localization and comprehensive feature capture by working at an optimal level of granularity (patches rather than pixels or regions in their entirety).

All these findings are relevant to the more general field of reliable AI. Choosing the most relevant features yields better interpretability and, more importantly, increases model robustness by focusing attention on patterns which contribute actively towards the task, rather than on circumstantial correlations. In use cases which are decision-critical (medicine) or safety-critical (self-driving cars), this XAI trait is particularly significant because making a model more comprehensible is as valuable as making it more accurate.

\mupax's main \textbf{limitation} is its computational cost. Its perturbation-based approach is more computationally demanding than gradient-based methods like GradCAM. This is a familiar trade-off between the quality of an explanation and its computational expense. Future work will investigate approximations that can reduce overall compute time while maintaining the level of explanation.

\section{Conclusion and Future Work}
\label{sec:conclusion}

Since \mupax's explainability theory  uses concepts from measure theory, it is deterministic, which is an essential quality in XAI. It is also generic in the sense it is model-- and loss--agnostic.
The most notable advancement in \mupax\ is its feature attribution with mathematically guaranteed convergence, addressed in Section~\ref{thm:evidence-general}. Moreover, this holds in any dimension space. 
Its robust feature selection through marking out irrelevant areas increases the overall explanation fidelity.
Alongside the explanation, 
the method also exhibits an increase in task performance across all tested data modalities.

In each test, the  frozen model (for 1D, 2D, 3D and Landmarks) was benchmarked using inputs filtered by various XAI techniques (including GradCAM, Integrated Gradients, SHAP, and LIME), \mupax's filtered output consistently achieved superior performance across various relevant metrics for all signal modalities.

Separately, \mupax\ also fills the gap in practical XAI methods for landmark detection, thereby branching the XAI research in that direction. It is worth mentioning that in a future research we will investigate the localization links between landmarks present in the retained area, with the goal of generating natural language descriptions of the explainations.

Future work will focus on creating computationally effective approximations to reduce perturbation overhead, extending the framework to temporal sequences and graph data, and incorporating human feedback for iterative tuning.  Values which are hard-wired in the current study will be parameterized, with further experiments towards fine-tuning these parameters.

It is also important to investigate how the output of \mupax\ is influenced by different model architectures.
Additionally, other data acquisition modalities not mentioned here, as well as applications other than healthcare, are worth exploring in their own right. 
A notable omission from this study is the task of semantic segmentation. Its absence is due to its dense per‐pixel output and overlap‐based evaluation metrics (IoU, Dice) resulting in strongly discontinuous retained regions. These would likely cause the computational costs to increase by at least an order of magnitude. Further work will investigate other ways of tackling segmentation. 

\newpage

\bibliographystyle{plainnat} 
\bibliography{my_bib}

\newpage

\section*{Appendix A}
\subsection*{\mupax\ Theorem Proof}  

\paragraph{Setup.}
Let $\mathbf{X}$, $f$, $\mu(\mathbf{X}_{\text{arg}})$, $\mathcal{U}$, $W$, and $p_W > 0$ be defined as in the theorem statement.
The sampling method generates selection vectors $s_i \sim \mathcal{U}$ and accepts them if $\mu(\mathbf{X}^{s_i}) \le W$. This is equivalent to rejection sampling.
Let $\{\overline{\mathbf{X}}^1, \dots, \overline{\mathbf{X}}^n\}$ be the sequence of $n$ masked inputs obtained from the first $n$ accepted selection vectors. As established by rejection sampling theory, these samples are i.i.d.\ according to the conditional distribution $\mathcal{D}_W$ (the distribution of $\mathbf{X}^s$ given $s \sim \mathcal{U}$ and $\mu(\mathbf{X}^s) \le W$).
Define the random variables $X_c(\alpha) = \mu'_c\,\overline{\mathbf{X}}^c(\alpha)$, where $\mu'_c = 1 / (\mu(\overline{\mathbf{X}}^c) + 1)$. The empirical average is $\chi_n(\alpha) = \frac{1}{n}\sum_{c=1}^n X_c(\alpha)$.
The target limit is the expectation under $\mathcal{D}_W$: $\chi(\alpha) = \mathbb{E}_{\mathbf{X}' \sim \mathcal{D}_W}[\mu'(\mathbf{X}')\,\mathbf{X}'(\alpha)]$.

\subsubsection*{Step 1: Boundedness}

Since $\mathbf{X}(\alpha) \ge 0$ by assumption, and $\overline{\mathbf{X}}^c(\alpha)$ is either $0$ (if $\alpha$ is masked) or $\mathbf{X}(\alpha)$ (if $\alpha$ is retained), it is possible to get $\overline{\mathbf{X}}^c(\alpha) \in [0, \mathbf{X}(\alpha)]$.
Also, $\mu(\overline{\mathbf{X}}^c) \ge 0$ implies $\mu_c + 1 \ge 1$, so $0 < \mu'_c = 1/(\mu_c+1) \le 1$.

Therefore, each term $X_c(\alpha)$ is bounded:
$$
0 \;\le\; X_c(\alpha) = \mu'_c\,\overline{\mathbf{X}}^c(\alpha) \;\le\; 1 \cdot \mathbf{X}(\alpha) = \mathbf{X}(\alpha).
$$
The average $\chi_n(\alpha)$ is also bounded by $\mathbf{X}(\alpha)$. Boundedness implies that the expectation $\chi(\alpha)$ and the variance $\sigma^2_\alpha$ (defined below) are finite.

\subsubsection*{Step 2: Convergence with Probability that tends to 1}
The random variables $\{X_c(\alpha)\}_{c=1}^n$ are i.i.d.\  because they are functions of the i.i.d.\ samples $\{\overline{\mathbf{X}}^c\}_{c=1}^n$ drawn from $\mathcal{D}_W$.
They have a finite mean $\mathbb{E}[X_c(\alpha)] = \mathbb{E}_{\mathbf{X}' \sim \mathcal{D}_W}[\mu'(\mathbf{X}')\,\mathbf{X}'(\alpha)] = \chi(\alpha)$.
Since the variables $X_c(\alpha)$ are i.i.d.\ and have a finite mean (implied by boundedness in Step 1), the Strong Law of Large Numbers (SLLN) applies directly:
$$
\chi_n(\alpha) \;=\; \frac{1}{n} \sum_{c=1}^n X_c(\alpha)
\;\xrightarrow{\text{a.s.}}\;
\mathbb{E}[X_c(\alpha)] \;=\; \chi(\alpha) \quad \text{as } n \to \infty.
$$
This establishes the almost sure pointwise convergence.

For the decomposition: Let $I(\alpha, \mathbf{X}')$ be an indicator variable that is 1 if $\alpha$ is retained in $\mathbf{X}'$ and 0 otherwise. Then $\mathbf{X}'(\alpha) = \mathbf{X}(\alpha) I(\alpha, \mathbf{X}')$.
\begin{align*} \chi(\alpha) &= \mathbb{E}_{\mathbf{X}' \sim \mathcal{D}_W}[\mu'(\mathbf{X}')\,\mathbf{X}(\alpha) I(\alpha, \mathbf{X}')] \\ &= \mathbf{X}(\alpha) \mathbb{E}_{\mathbf{X}' \sim \mathcal{D}_W}[\mu'(\mathbf{X}')\, I(\alpha, \mathbf{X}')] \\ &= \mathbf{X}(\alpha) \mathbb{E}_{\mathbf{X}' \sim \mathcal{D}_W}[\mu'(\mathbf{X}') \,|\, I(\alpha, \mathbf{X}')=1] \, \mathbb{P}_{\mathbf{X}' \sim \mathcal{D}_W}(I(\alpha, \mathbf{X}')=1) \\ &= \mathbf{X}(\alpha)\,\mathbb{E}_{\mathbf{X}' \sim \mathcal{D}_W}\bigl[\mu'(\mathbf{X}') \,\mid \alpha \text{ retained in } \mathbf{X}'\bigr]\, \mathbb{P}_{\mathbf{X}' \sim \mathcal{D}_W}\bigl(\alpha \text{ retained in } \mathbf{X}'\bigr). \end{align*}
This confirms the decomposition formula, where the expectation and probability are taken with respect to the conditional distribution $\mathcal{D}_W$. Let $A'(\alpha) = \mathbb{P}_{\mathbf{X}' \sim \mathcal{D}_W}(\alpha \text{ retained in } \mathbf{X}')$.

\subsubsection*{Step 3: Central Limit Theorem for Convergence Rate}
Since $X_c(\alpha)$ are i.i.d.\ with mean $\mu_\alpha = \chi(\alpha)$ and are bounded (Step 1), their variance $\sigma^2_\alpha = \mathbb{E}_{\mathbf{X}' \sim \mathcal{D}_W}\bigl[X_c(\alpha)^2\bigr] - \mu_\alpha^2$ is finite. For example, $\mathbb{E}_{\mathbf{X}' \sim \mathcal{D}_W}[X_c(\alpha)^2] \le \mathbf{X}(\alpha)^2$, so $\sigma^2_\alpha \le \mathbf{X}(\alpha)^2$.
By the Lindeberg-Lévy Central Limit Theorem (CLT), the distribution of the scaled difference between the sample mean and the true mean converges to a normal distribution:
$$
\sqrt{n}\,\bigl(\chi_n(\alpha)-\chi(\alpha)\bigr)
\;\xrightarrow{d}\;
\mathcal{N}\bigl(0,\sigma^2_\alpha\bigr) \quad \text{as } n \to \infty.
$$
This indicates that the convergence rate of the sample mean $\chi_n(\alpha)$ to the true mean $\chi(\alpha)$ is typically of the order $O_p(n^{-1/2})$.

\medskip
Given the boundedness of the terms (Step 1), the establishment of almost sure convergence via the SLLN justified by the i.i.d.\ nature of samples from $\mathcal{D}_W$ obtained through rejection sampling (Step 2), and the characterization of the convergence rate via the CLT (Step 3), the proof is complete.
\newpage

\section*{Appendix B}
 
\subsection*{Ablation Studies}  

\begin{table}[ht]
    \centering
    \footnotesize
    \caption{Ablation Study Results: Impact of Hyperparameter Variation on Masked Input Performance (Macro F1-Score) for the 2D Cat vs.\ Dog Classification Task.}
    \label{tab:ablation_results}
    \begin{tabular}{llllr}
        \toprule
        \textbf{Method} & \textbf{Parameter Varied} & \textbf{Condition} & \textbf{Value} & \textbf{Macro F1} \\
        \midrule
        \multirow{4}{*}{\mupax} 
        & \textit{Baseline} & - & - & 0.95 \\ 
        & Num Samples (K) & Lower & 500 & 0.94 \\
        & Chunk Size (pixels) & Coarser & 16x16 & 0.86 \\
        & Selection Threshold & Less Strict & 0.5 & 0.71 \\
        \midrule
        \multirow{4}{*}{LIME} 
        & \textit{Baseline} & - & - & 0.90 \\ 
        & Num Samples & Fewer & 100 & 0.88 \\
        & Num Features & More & 15 & 0.92 \\
        & Feature Selection & Include Negative & False & 0.82 \\
        \midrule
        \multirow{5}{*}{Grad-CAM} 
        & \textit{Baseline} & - & - & 0.47 \\ 
        & Conv Layer & Earlier & Stage 2 & 0.48 \\
        & Mask Threshold & Higher & 0.8 & 0.38 \\
        & Mask Threshold & Lower & 0.2 & 0.45 \\
        & Gradient Pooling & Max & Max & 0.33 \\
        \midrule
        \multirow{4}{*}{SHAP (GradientExplainer)} 
        & \textit{Baseline} & - & - & 0.45 \\ 
        & Background Samples & Single Black & 1 & 0.33 \\
        & Background Samples & Fewer & 10 & 0.33 \\
        & Mask Threshold & Higher & 0.5 & 0.33 \\
        \midrule
        \multirow{4}{*}{Integrated Gradients (IG)} 
        & \textit{Baseline} & - & - & 0.66 \\ 
        & Integration Steps & More & 100 & 0.59 \\
        & Baseline & Average Image & Avg & 0.56 \\
        & Masking Percentage & Stricter & Top 1\% & 0.54 \\
        \bottomrule
    \end{tabular}
\end{table}

\subsection{Discussion of Results in Ablation Studies}
\label{sec:ablation}

To assert the sensitivity of the experimented XAI methods to their respective hyperparameters, ablation studies were conducted. For each of the methods (\mupax, LIME, Grad-CAM, SHAP GradientExplainer, and Integrated Gradients), a baseline configuration akin to the main experiments (Section \ref{sec:experiments}) was established. Key hyperparameters were modified systematically and the impact on the model performance with the input mask was measured in terms of Macro F1-score on the 2D image classification task. Results of the ablation study are reported in Table~\ref{tab:ablation_results}.
These results indicate the sensitivity of the various XAI methods to the respective component.

For \mupax, performance was kept high (Macro F1 = 0.94) even when the number of perturbation samples (`K`) was reduced by an order of magnitude (50,000 to 5,000), which is an indication of some insensitivity to sampling density. Raising the chunk size, however (from 8x8 to 16x16), resulted in the decrease in performance being more significant (Macro F1 = 0.86), which is an indication that explanation granularity impacts the ability to accurately predict relevant features. Furthermore, significantly relaxing the loss threshold setting it to 0.5 instead of the original, 0.8 (which means setting the parameter $W = 1.0-threshold = 1.0-0.8= 20\%$ as specified earlier) is equivalent to sampling based on a weaker model performance threshold which, in turn, resulted in a significant decrease in performance (Macro F1 = 0.71). This highlights the value of the sample selection strategy in preserving model performance on the derived explanation.

LIME results were robust. Reducing the sample size (from 500 to 100) resulted in a minor decrease in performance (Macro F1 = 0.88). Raising the number of features to use for the explanation (from 5 to 15) resulted in a minor increase in Macro F1 to 0.92, possibly by obtaining a fuller set of relevant superpixels, but the amount of retrieved image was high, i.e. almost the majority of the image was retrieved, including background, thus not focusing on the most important patterns (like dogs and cats). Removing the the positive only constraint, hence including superpixels that are negatively correlated with the prediction, resulted in a decrease in performance (Macro F1 = 0.82). Grad-CAM was very parameter-sensitive. Modifying the target convolutional layer to an earlier stage (Stage 2) instead of a later stage (Stage 3) did not affect the already poor baseline performance substantially. The modifications in the mask binarization threshold had a mixed effect: raising the threshold (0.8) reduced the performance significantly (Macro F1 = 0.38), while reducing the threshold (0.2) resulted in performances close to the baseline (Macro F1 = 0.45). Using max pooling rather than mean pooling over gradients resulted in very poor performance (Macro F1 = 0.33). These results point to Grad-CAM's sensitivity to proper layer choice and thresholding in order to generate effective masks.

The SHAP GradientExplainer version was highly sensitive to background distribution and threshold values. Both a single black image and a very limited background sample set (10) produced the same poor result (Macro F1 = 0.33). Similarly, an increased binarization threshold (0.5) also produced the same low score. This indicates that the performance of SHAP GradientExplainer is highly sensitive to the need for a representative background dataset and careful threshold value for mask generation.

Integrated Gradients also performed poorly on the majority of the baseline variations. Increasing the number of integration steps (from 25 to 100) lowered performance (Macro F1 = 0.59). Using an average image as the baseline instead of a zero baseline also fared worse (Macro F1 = 0.56). Forcing the mask by considering only the top 1\% of attributed pixels lowered the score (Macro F1 = 0.54). 

In general, the ablation experiments indicate that while all methods exhibit some parameter sensitivity, \mupax\ and LIME were comparatively more stable to variations than Grad-CAM, SHAP GradientExplainer, and Integrated Gradients, which were more unstable or exhibited degraded baseline performance in this masked test setup. \mupax's sensitivity to chunk size and selection threshold provide insights into granularity of explanations vs.\ computational cost (through sample size) vs.\ fidelity trade-offs. 
In practice, it is necessary to find trade-offs. Chunk size is a trade-off between feature scale and expense (large chunk) and granularity (small chunk), depending on data modality. Threshold $W$ is a trade-off between efficiency ($p_W$) and strictness. Stricter $W$ (low percentile) produces crisper explanations but at the cost of more samples ($\N{total}$). Less strict $W$ (high percentile) is cheaper but may water down results. The suggestion is to start moderately (e.g.\ $20^{th}$ percentile) and adjust based on observed quality and available computational resources.

\end{document}